\documentclass[sigconf]{aamas}  

\usepackage{booktabs}
\usepackage{todonotes}

\newcommand{\wrt}{\emph{wrt}}

\usepackage{graphicx}
\usepackage{tikz}

\usepackage{pgfplots}
\pgfplotsset{compat=1.14}
\usepgfplotslibrary{colorbrewer}

\pgfplotsset{
    /pgfplots/ybar legend/.style={
    /pgfplots/legend image code/.code={%
       \draw[##1,/tikz/.cd,yshift=-0.25em]
        (0cm,0cm) rectangle (3pt,0.6em);},
   },
}

\usepackage{mleftright,xparse}
\usepackage{mathtools}

\NewDocumentCommand\xDeclarePairedDelimiter{mmm}
 {%
  \NewDocumentCommand#1{som}{%
   \IfNoValueTF{##2}
    {\IfBooleanTF{##1}{#2##3#3}{\mleft#2##3\mright#3}}
    {\mathopen{##2#2}##3\mathclose{##2#3}}%
  }%
 }

\xDeclarePairedDelimiter{\pa}{(}{)}

\xDeclarePairedDelimiter{\interval}{[}{]}
\xDeclarePairedDelimiter{\card}{\vert}{\vert}
\xDeclarePairedDelimiter{\norm}{\lVert}{\rVert}
\xDeclarePairedDelimiter{\tuple}{\langle}{\rangle}

\usepackage{flushend}
\setcopyright{ifaamas}  
\acmDOI{doi}  
\acmISBN{}  
\acmConference[AAMAS'20]{Proc.\@ of the 19th International Conference on Autonomous Agents and Multiagent Systems (AAMAS 2020), B.~An, N.~Yorke-Smith, A.~El~Fallah~Seghrouchni, G.~Sukthankar (eds.)}{May 2020}{Auckland, New Zealand}  
\acmYear{2020}  
\copyrightyear{2020}  
\acmPrice{}  

\begin{document}

\title{Predicting Requests in Large-Scale Online P2P Ridesharing}  

\author{Filippo Bistaffa}
\affiliation{%
  \institution{IIIA-CSIC}
  \streetaddress{Campus UAB}
  \city{Barcelona}
  \state{Spain}
  \postcode{08193}
}
\email{filippo.bistaffa@iiia.csic.es}

\author{Juan A. Rodr\'{i}guez-Aguilar}
\affiliation{%
  \institution{IIIA-CSIC}
  \streetaddress{Campus UAB}
  \city{Barcelona}
  \state{Spain}
  \postcode{08193}
}
\email{jar@iiia.csic.es}

\author{Jes\'{u}s Cerquides}
\affiliation{%
  \institution{IIIA-CSIC}
  \streetaddress{Campus UAB}
  \city{Barcelona}
  \state{Spain}
  \postcode{08193}
}
\email{cerquide@iiia.csic.es}

\begin{abstract}  
\emph{Peer-to-peer ridesharing} (P2P-RS) enables people to arrange one-time rides with their own private cars, without the involvement of professional drivers. It is a prominent \emph{collective intelligence} application producing significant benefits both for individuals (reduced costs) and for the entire community (reduced pollution and traffic), as we showed in a recent publication where we proposed an online approximate solution algorithm for large-scale P2P-RS.

In this paper we tackle the fundamental question of assessing the benefit of predicting ridesharing requests in the context of P2P-RS optimisation. 
Results on a public real-world show that, by employing a \emph{perfect} predictor, the total reward can be improved by $5.27\%$ with a forecast horizon of $1$ minute.
On the other hand, a vanilla \emph{long short-term memory} neural network cannot improve upon a baseline predictor that simply replicates the previous day's requests, whilst achieving an almost-double accuracy.
\end{abstract}


\maketitle


\section{Introduction}

With the growing popularity of the sharing economy, ridesharing services are called to transform urban mobility. Shared mobility is expected to have major environmental and economic impacts by reducing pollution, traffic congestion, and energy consumption, which could be further enhanced with the advent of electrical vehicles. Moreover, ridesharing is set to become even more attractive in a near-future world of self-driving cars, spurring a transition from solo driving to mass transit.
This major endeavour addresses key topics in Sustainable Development Goals 11 (``sustainable cities and communities'') and 7 (``affordable and clean energy'').

The concept of ridesharing has recently received increasing attention, both in the transportation industry (e.g., UberPool, Lyft, and Maramoja) and in academia. However, \emph{ridesharing} may have different interpretations that result in significant differences in the computational models that can be used to represent and address the problem. 
This problem has been addressed from a computational view point only very recently using various algorithms \cite[and references therein]{alonso2017demand,tits}. Most of these approaches (e.g., \cite{alonso2017demand}) focus on maximising the benefits of the passengers considering mainly the concept of quality of service (i.e., minimising the delay experienced by the users). In a recent work~\cite{tits} we proposed a different perspective on the \emph{peer-to-peer ridesharing} (P2P-RS) problem by considering not only the quality of service, but also the environmental benefits resulting from ridesharing (e.g., CO\textsubscript{2} and traffic congestion reduction). This perspective address the trade-off between those two objectives when forming shared rides, looking at ridesharing as a method to foster sustainable mobility for policy-makers and not only as a profitable alternative for commuters.

On the one hand, despite tackling an inherently \emph{online} problem (i.e., ridesharing requests are not known in advance), the majority of the works in the ridesharing optimisation literature do not employ \emph{predictions} of forthcoming requests to improve the quality of their online solutions.\footnote{In \cite{tits} we propose a \emph{look-ahead} mechanism that filters out immediately profitable cars whose formation could hinder the formation of even better cars in the near future.} 
On the other hand, even though there exist some works aiming at predicting \emph{mobility on demand} requests (see, e.g., \cite{8215650}), such works only look at the \emph{accuracy} of the predictions,
whereas their impact on the optimisation process remains unclear.

In this paper we tackle the fundamental question of assessing the benefit of predictions in the context of P2P-RS optimisation. Specifically, we present our preliminary study that aims at including predictions in our P2P-RS solution technique~\cite{tits}, with the objective of improving the total reward of the computed solution.
Results on a public real-world  dataset \cite{taxinyc} show that, by employing a \emph{perfect} predictor, the total reward can be improved by $5.27\%$ ($6.31\%$ during weekend days) with a forecast horizon of $1$ minute.
On the other hand, a \emph{long short-term memory} (LSTM) neural network~\cite{hochreiter1997long} cannot improve upon a predictor that simply replicates the previous day's requests, whilst achieving an almost-double accuracy.

\section{The P2P ridesharing problem}

We consider the P2P-RS problem as defined in \cite{tits}, i.e., as an \emph{online stochastic scheduling} problem~\cite{hentenryck2009online} over a discrete time horizon $H=[1,h]$.
Our problem takes place in an area divided in $n$ zones, i.e., $Z=\{1,\ldots,n\}.$
At each step $t\in H$, the system receives a (possibly empty) set of requests $R_t$.
Each request $r\in R_t$ is a tuple $\tuple{i,j,d,\delta}$ characterised by a starting zone $i\in Z$ and a destination zone $j\in Z$, a Boolean value $d\in\mathbb B$ indicating whether the corresponding commuter has a car or not (i.e., whether it is a driver or not), and the maximum time $\delta\in H$ the commuter is willing to wait to be assigned to a car.
Formally, $r\in\mathcal R=Z\times Z\times\mathbb B\times H.$
Hence, $\tuple{R_1,\ldots,R_t,\ldots,R_h}$ represents the input of the problem.
We measure the performance of a given set $S$ of requests (i.e., a car) both in terms of \emph{environmental benefits} and \emph{quality of service} (QoS). Formally, we define $V : 2^{\mathcal R}\to\mathbb R$ as the total reward (i.e., considering environmental benefits and the quality of service) associated to a given car $S$ as
\begin{multline}\label{eq:v}
V\pa{S}=
\rho\textsubscript{CO\textsubscript{2}}\cdot E\textsubscript{CO\textsubscript{2}}\pa{S}
+ \rho\textsubscript{noise} \cdot E\textsubscript{noise}\pa{S} + \\
\rho\textsubscript{traffic} \cdot E\textsubscript{traffic}\pa{S} + \rho\textsubscript{QoS} \cdot Q\pa{S},
\end{multline}
where $\rho\textsubscript{CO\textsubscript{2}}$, $\rho\textsubscript{noise}$, and $\rho\textsubscript{traffic}$ measure the importance of the corresponding environmental benefit and $\rho\textsubscript{QoS}$ that of the QoS.
We refer the interested reader to \cite{tits} for further technical details.

In our previous work~\cite{tits}, we defined $Q : 2^{\mathcal R}\to\mathbb R_{\leq 0}$, namely the \emph{quality of service} associated to a given car $S$, as $Q\pa{S}=-\sum_{r\in S}\frac{t_r-t^*_r}{t_r},$
where $t_r$ and $t^*_r$ are the travel times associated to request $r$ with and without ridesharing, respectively.
Since $t^*_r$ is the optimal travel time associated to $r=\tuple{i,j,d,\delta}$ (i.e., obtained by driving through the shortest path from zone $i$ to zone $j$), then $0\leq\frac{t_r-t^*_r}{t_r}\leq 1$.

In this paper we extend such a definition by considering a more realistic QoS measure that also takes into account the \emph{to-be-assigned} (TBA) delay, i.e., the delay from the time $t_r$ of arrival of a request $r$ and the time of its assignment to a car $S$. Notice that, given a car $S$, the assignment time of any request $r\in S$ to $S$ corresponds to the time when $S$ is actually formed, i.e., when the the latest request $r\in S$ has entered the system. Formally,
\begin{equation}\label{eq:qos_tba}
Q\pa{S}=-\sum_{r\in S}\frac{t_r-t^*_r}{t_r}-\sum_{r\in S}\frac{\max_{r'\in S}t_{r'}-t_r}{\delta_r}.
\end{equation}
Notice that, in order to be consistent with our previous definition of $Q\pa{\cdot}$, we normalise each TBA delay by dividing it by the maximum assignment time, hence ensuring that $0\leq\frac{\max_{r'\in S}t_{r'}-t_r}{\delta_r}\leq 1$.

Intuitively, the P2P-RS problem aims at arranging, at each time step $t$, a (possibly empty) set $\mathcal S_t$ of non-overlapping cars among the current set of active requests, with the objective of maximising the sum of the associated rewards over the entire time horizon $H$.
The formal definition of the P2P-RS problem can be found in \cite{tits}.

\subsection{Solving the P2P-RS Problem}

We tackle the P2P-RS problem by solving a \emph{sequence} of offline problems by means of our approximate offline approach~\cite{tits}.
Our approximate offline approach employs a well-known technique for approximately solving time-constrained large-scale combinatorial optimisation problems modelled as \emph{integer linear programs} (ILP), which consists in (1) removing those variables from the model that, a priori, do not seem to help for the generation of good solutions, and (2) passing the reduced ILP to a solver in order to get the best solution possible in the available computation time.

In the specific case of our approximate offline approach, we iteratively apply a probabilistic greedy heuristic to generate feasible cars that represent ``good candidates'' for the final optimisation solution.
We then formulate and solve an ILP (specifically, a \emph{weighted set packing} problem) whose decision variables correspond only to the cars in the above-mentioned set of good candidates, rather than to the set of all possible cars (whose enumeration would be infeasible in the time budget we consider for the P2P-RS scenario, i.e., 1 minute for solving the offline problems corresponding to each time step).
Finally, the set of cars resulting from the solution of the above-discussed reduced ILP is filtered by means of a \emph{look-ahead} mechanism that discards immediately profitable cars whose formation could hinder the formation of even better cars in the near future.
Notice that, even though such a technique allows us to put each offline solution in the context of the overall online problem, it does not represent a \emph{predictive} technique, since it does not take into account the history of requests, and, more importantly, it is applied \emph{after} the optimisation, and not \emph{before}.

\section{Predictions for P2P ridesharing}

As previously discussed, none of the recent works on ridesharing optimisation \cite[and references therein]{alonso2017demand,tits} employs prediction to improve the quality of the solution in terms of total reward.
Indeed, rather than employing a \emph{predictive} approach, the majority of the works embrace  \emph{reactive} techniques that aim at taking into account the requests in the optimisation process once they already materialised inside the system (including our own \emph{look-ahead} method~\cite{tits}).
As a possible notable exception, Bicocchi \emph{et al.}~\cite{bicocchi2017recommending} proposed a systems for recommending ridesharing matches based on the historical data provided by a large Italian telecom operator.

To the best of our knowledge, the majority of the works focusing on predicting \emph{mobility on demand} requests are usually interested in measuring and maximising the accuracy of predicting the number of requests (e.g., \cite{8215650}).
On the other hand, the actual impact of such predictions on the optimisation process remains unclear.

Here we tackle the following fundamental questions: are predictions beneficial \wrt{} to reactive techniques in the context of P2P-RS optimisation? If so, which is the best \emph{forecast horizon} $f$? Which is the maximum improvement in terms of total reward that one can expect by employing predictions? How do predictions impact on the solution quality of our approximate offline approach, which receives a larger amount of input requests in the same time budget?

To address the above questions, we incorporate 3 different types of predictors in our P2P-RS online solution algorithm: (1) a \emph{perfect} predictor with 100\% accuracy, (2) a predictor that replicates the requests of the previous day (as also considered in \cite{8215650}), and (3) a vanilla LSTM neural network~\cite{hochreiter1997long}, comprised of a LSTM layer with a hidden vector $h$ of size $Z\times Z$ (i.e., the number of different possible requests), a fully connected layer, and a ReLU layer.

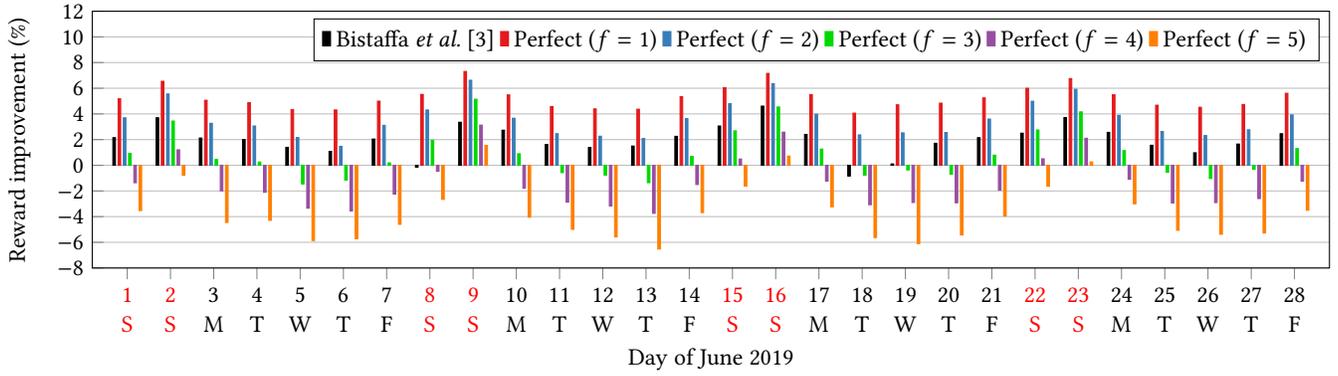
\begin{figure*}[t]
\centering
\begin{tikzpicture}
    \begin{axis}[
        x post scale=2.4,
        y post scale=0.6,
    	xlabel={Day of June 2019},
    	ylabel={Reward improvement (\%)},
    	y label style={at={(axis description cs:-0.075,0.5)},anchor=north},
        ymin=-8,
        ymax=12,
        xtick=data,
        xticklabels={
            \textcolor{red}{1}\\\textcolor{red}{S},
            \textcolor{red}{2}\\\textcolor{red}{S},
            3\\M,
            4\\T,
            5\\W,
            6\\T,
            7\\F,
            \textcolor{red}{8}\\\textcolor{red}{S},
            \textcolor{red}{9}\\\textcolor{red}{S},
            10\\M,
            11\\T,
            12\\W,
            13\\T,
            14\\F,
            \textcolor{red}{15}\\\textcolor{red}{S},
            \textcolor{red}{16}\\\textcolor{red}{S},
            17\\M,
            18\\T,
            19\\W,
            20\\T,
            21\\F,
            \textcolor{red}{22}\\\textcolor{red}{S},
            \textcolor{red}{23}\\\textcolor{red}{S},
            24\\M,
            25\\T,
            26\\W,
            27\\T,
            28\\F
        },
        xticklabel style={align=center},
        ytick distance={2},
        xtick pos=left,
	    ytick pos=left,
	    ymajorgrids,
	    enlarge x limits=0.03,
        bar width=0.06,
        ybar=1pt,
        scaled y ticks=false,
        /pgf/number format/set thousands separator={},
        legend style={
            at={(0.992,0.972)},
            anchor=north east,
        },
        legend entries={
            Bistaffa \emph{et al.}~\cite{tits},
            Perfect ($f=1$),
            Perfect ($f=2$),
            Perfect ($f=3$),
            Perfect ($f=4$),
            Perfect ($f=5$),
        },
        legend cell align={left},
        legend columns=6,
        table/col sep=comma,
        cycle list/Set1,
        every axis plot/.append style={fill}
    ]
    
    \addplot [black] table [x index=0, y index=1] {rw_impr.csv};
    \addplot [Set1-A] table [x index=0, y index=2] {rw_impr.csv};
    \addplot [Set1-B] table [x index=0, y index=3] {rw_impr.csv};
    \addplot [black!15!green] table [x index=0, y index=4] {rw_impr.csv};
    \addplot [Set1-D] table [x index=0, y index=5] {rw_impr.csv};
    \addplot [Set1-E] table [x index=0, y index=6] {rw_impr.csv};

    \end{axis}
\end{tikzpicture}
\caption{\label{fig:perf_impr}Total reward percentage improvement \wrt{} no predictions (best viewed in colour).}
\end{figure*}

\begin{figure*}[t]
\centering
\begin{tikzpicture}
    \begin{axis}[
        x post scale=2.4,
        y post scale=0.6,
    	xlabel={Day of June 2019},
    	ylabel={Average pool size (requests)},
    	y label style={at={(axis description cs:-0.075,0.5)},anchor=north},
        ymin=200,
        ymax=1600,
        xtick=data,
        xticklabels={
            \textcolor{red}{1}\\\textcolor{red}{S},
            \textcolor{red}{2}\\\textcolor{red}{S},
            3\\M,
            4\\T,
            5\\W,
            6\\T,
            7\\F,
            \textcolor{red}{8}\\\textcolor{red}{S},
            \textcolor{red}{9}\\\textcolor{red}{S},
            10\\M,
            11\\T,
            12\\W,
            13\\T,
            14\\F,
            \textcolor{red}{15}\\\textcolor{red}{S},
            \textcolor{red}{16}\\\textcolor{red}{S},
            17\\M,
            18\\T,
            19\\W,
            20\\T,
            21\\F,
            \textcolor{red}{22}\\\textcolor{red}{S},
            \textcolor{red}{23}\\\textcolor{red}{S},
            24\\M,
            25\\T,
            26\\W,
            27\\T,
            28\\F
        },
        xticklabel style={align=center},
        ytick distance={200},
        xtick pos=left,
	    ytick pos=left,
	    ymajorgrids,
	    enlarge x limits=0.03,
        bar width=0.06,
        ybar=1pt,
        scaled y ticks=false,
        /pgf/number format/set thousands separator={},
        legend style={
            at={(0.992,0.028)},
            anchor=south east,
        },
        legend entries={
            Bistaffa \emph{et al.}~\cite{tits},
            Perfect ($f=1$),
            Perfect ($f=2$),
            Perfect ($f=3$),
            Perfect ($f=4$),
            Perfect ($f=5$),
        },
        legend cell align={left},
        legend columns=6,
        table/col sep=comma,
        cycle list/Set1,
        every axis plot/.append style={fill}
    ]
    
    \addplot [black] table [x index=0, y index=1] {pool_size.csv};
    \addplot [Set1-A] table [x index=0, y index=2] {pool_size.csv};
    \addplot [Set1-B] table [x index=0, y index=3] {pool_size.csv};
    \addplot [black!15!green] table [x index=0, y index=4] {pool_size.csv};
    \addplot [Set1-D] table [x index=0, y index=5] {pool_size.csv};
    \addplot [Set1-E] table [x index=0, y index=6] {pool_size.csv};

    \end{axis}
\end{tikzpicture}
\caption{\label{fig:pool_size}Average number of requests (including predictions) in the pool (best viewed in colour).}
\end{figure*}
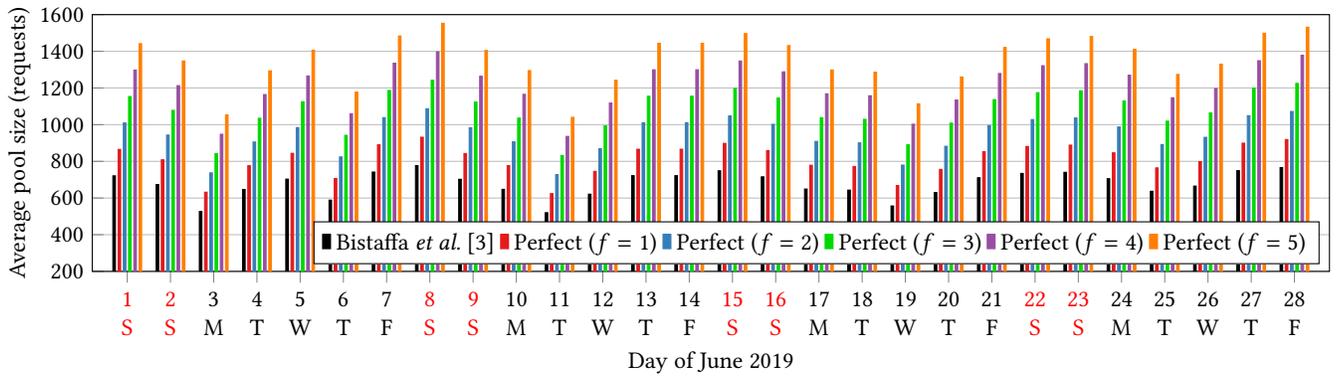

\section{Experimental Evaluation}
\subsection{Experimental Methodology}

Our experimental evaluation has the following objectives:
\begin{enumerate}
    \item determine the maximum improvement in terms of total reward when employing a \emph{perfect} predictor,
    \item determine the best forecast horizon $f$ for predictions,
    \item compare a vanilla LSTM predictor with a \emph{perfect} predictor, a \emph{yesterday} predictor, and our \emph{look-ahead} approach~\cite{tits}.
\end{enumerate}
If not otherwise specified, our experimental evaluation follows the same methodology adopted in our previous work~\cite{tits}.
We employ the same publicly available real-world dataset of taxi trips in Manhattan, New York City~\cite{taxinyc}, considering the most recent data available at the time of writing (i.e., June 2019).
As previously mentioned, in all our experiments we consider a more realistic measure for QoS \wrt{} \cite{tits}, i.e., one that also takes into account the \emph{to-be-assigned} delay (Equation~\ref{eq:qos_tba}).
In all our experiments we consider a \emph{maximum assignment time} $\delta=5$ and a time horizon $h=1$ day.

Following \cite{8215650}, we employ the \emph{symmetric mean absolute percentage error} (SMAPE) \cite{FLORES198693} to measure the accuracy of all predictors:
$$\text{SMAPE} = \frac 1 n \sum_{t=1}^n \frac{\left|P_t-G_t\right|}{(P_t+G_t)/2},$$
where $P_t$ and $G_t$ are the prediction and the ground truth at time $t$.

Our LSTM predictor 
has been trained on a machine with a 12-cores 3.70GHz CPU, 32GB of RAM, and two GeForce RTX 2080 Ti GPUs.
All optimisation experiments have been run on a cluster whose computing nodes have two 4-cores 2.26GHz CPUs and 32GB of RAM, using CPLEX 12.7 as ILP solver.\footnote{Source code: \url{https://github.com/filippobistaffa/P2P-RS/tree/OptLearnMAS}.}

\subsection{Perfect Predictor}

First, we aim at determining the maximum improvement in terms of total reward by employing a \emph{perfect} predictor when considering a forecast horizon $f$ up to the considered maximum assignment time $\delta=5$, i.e., $f\in\{1,2,3,4,5\}$.
Notice that, since the behaviour and the performance of our online optimisation algorithm depends on the parameters $\gamma$, $d_{\text{rate}}$, and $l_{\text{size}}$ (namely, the solution time budget, the determinism rate, and the candidate list size), following \cite{tits} we determined the values of such parameters for each $f$ using the \texttt{irace} package.
Figure \ref{fig:perf_impr} shows the total reward percentage improvement when using a perfect predictor as well as our \emph{look-ahead} approach~\cite{tits}, compared with the case with no prediction (i.e., $f=0$).
Results clearly show two interesting behaviours.

First, for every considered prediction technique, the improvement obtained by employing predictions follows a pattern that clearly repeats every week, reaching the maximum improvement during the weekend (days marked in red).
Second, results clearly show that the reward improvement decreases when increasing the forecast horizon $f$, providing a tangible benefit only up to $f=2$.
Although the explanation of these results is not straightforward, by looking at the average number of requests in the pool in the corresponding days (Figure \ref{fig:pool_size}) we get some interesting insights.

Indeed, we notice a correlation between the behaviour in Figure \ref{fig:perf_impr} and \ref{fig:pool_size}, suggesting that a slightly larger amount of requests during the weekend allows for a larger reward improvement due to the employment of predictions.
These observations seem related to our previous findings, as in \cite{tits} we noticed that the quality of our online approach \wrt{} the offline approach progressively increased as the problem size increased. 
Nonetheless, it seems that once the number of requests in the pool grows excessively due to the longer forecast horizon (the average pool size for $f=5$ is more than double \wrt{} the case without predictions, i.e., when using the \emph{look-ahead} approach), our approximate offline approach cannot produce a solution of the same quality in the same time budget (i.e., 1 minute). 
We will further investigate these aspects in our future work.

Our main conclusion is that our online optimisation algorithm clearly benefits from perfect predictions only when using a forecast horizon $f\in\{1,2\}$, reaching an average improvement of $5.27\%$ ($6.31\%$ during weekend days) with $f=1$.
In comparison, our \emph{look-ahead} approach achieves an average improvement of $2\%$. 

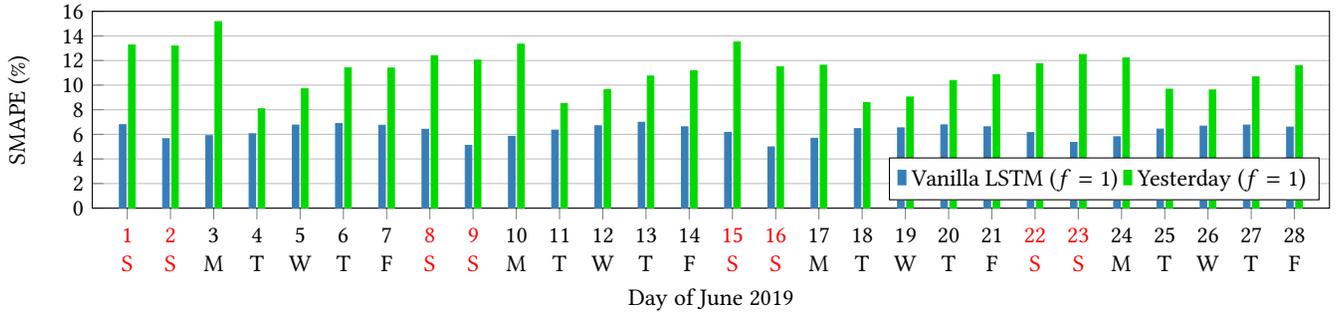
\begin{figure*}[t]
\centering
\begin{tikzpicture}
    \begin{axis}[
        x post scale=2.4,
        y post scale=0.46,
    	xlabel={Day of June 2019},
    	ylabel={SMAPE (\%)},
    	y label style={at={(axis description cs:-0.075,0.5)},anchor=north},
        ymin=0,
        ymax=16,
        xtick=data,
        xticklabels={
            \textcolor{red}{1}\\\textcolor{red}{S},
            \textcolor{red}{2}\\\textcolor{red}{S},
            3\\M,
            4\\T,
            5\\W,
            6\\T,
            7\\F,
            \textcolor{red}{8}\\\textcolor{red}{S},
            \textcolor{red}{9}\\\textcolor{red}{S},
            10\\M,
            11\\T,
            12\\W,
            13\\T,
            14\\F,
            \textcolor{red}{15}\\\textcolor{red}{S},
            \textcolor{red}{16}\\\textcolor{red}{S},
            17\\M,
            18\\T,
            19\\W,
            20\\T,
            21\\F,
            \textcolor{red}{22}\\\textcolor{red}{S},
            \textcolor{red}{23}\\\textcolor{red}{S},
            24\\M,
            25\\T,
            26\\W,
            27\\T,
            28\\F
        },
        xticklabel style={align=center},
        ytick distance={2},
        xtick pos=left,
	    ytick pos=left,
	    ymajorgrids,
	    enlarge x limits=0.03,
        bar width=0.15,
        ybar=1pt,
        scaled y ticks=false,
        /pgf/number format/set thousands separator={},
        legend style={
            at={(0.992,0.04)},
            anchor=south east,
        },
        legend entries={
            Vanilla LSTM ($f=1$),
            Yesterday ($f=1$),
        },
        legend cell align={left},
        legend columns=6,
        table/col sep=comma,
        cycle list/Set1,
        every axis plot/.append style={fill}
    ]
    
    \addplot [Set1-B] table [x index=0, y index=1] {smape.csv};
    \addplot [black!15!green] table [x index=0, y index=2] {smape.csv};

    \end{axis}
\end{tikzpicture}
\caption{\label{fig:smape}SMAPE for LSTM and \emph{yesterday} predictors (best viewed in colour).}
\end{figure*}

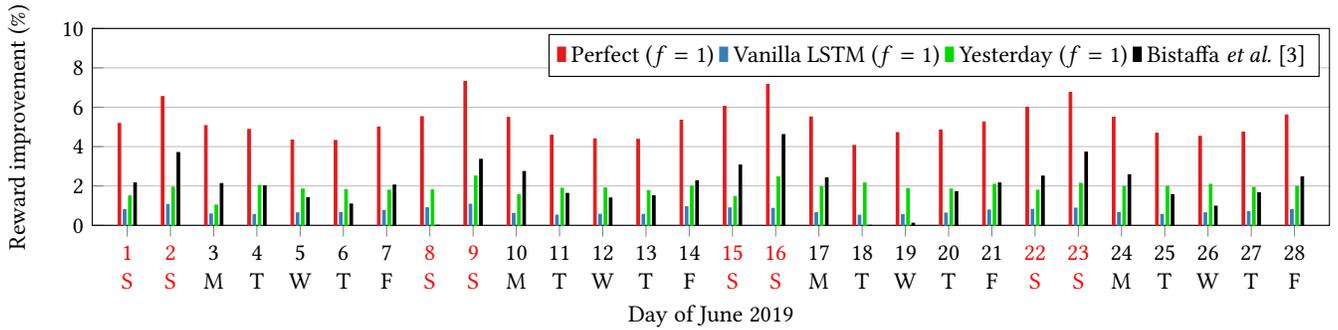
\begin{figure*}[t]
\centering
\begin{tikzpicture}
    \begin{axis}[
        x post scale=2.4,
        y post scale=0.46,
    	xlabel={Day of June 2019},
    	ylabel={Reward improvement (\%)},
    	y label style={at={(axis description cs:-0.075,0.5)},anchor=north},
        ymin=0,
        ymax=10,
        xtick=data,
        xticklabels={
            \textcolor{red}{1}\\\textcolor{red}{S},
            \textcolor{red}{2}\\\textcolor{red}{S},
            3\\M,
            4\\T,
            5\\W,
            6\\T,
            7\\F,
            \textcolor{red}{8}\\\textcolor{red}{S},
            \textcolor{red}{9}\\\textcolor{red}{S},
            10\\M,
            11\\T,
            12\\W,
            13\\T,
            14\\F,
            \textcolor{red}{15}\\\textcolor{red}{S},
            \textcolor{red}{16}\\\textcolor{red}{S},
            17\\M,
            18\\T,
            19\\W,
            20\\T,
            21\\F,
            \textcolor{red}{22}\\\textcolor{red}{S},
            \textcolor{red}{23}\\\textcolor{red}{S},
            24\\M,
            25\\T,
            26\\W,
            27\\T,
            28\\F
        },
        xticklabel style={align=center},
        ytick distance={2},
        xtick pos=left,
	    ytick pos=left,
	    ymajorgrids,
	    enlarge x limits=0.03,
        bar width=0.06,
        ybar=1pt,
        scaled y ticks=false,
        /pgf/number format/set thousands separator={},
        legend style={
            at={(0.992,0.972)},
            anchor=north east,
        },
        legend entries={
            Perfect ($f=1$),
            Vanilla LSTM ($f=1$),
            Yesterday ($f=1$),
            Bistaffa \emph{et al.}~\cite{tits},
        },
        legend cell align={left},
        legend columns=6,
        table/col sep=comma,
        cycle list/Set1,
        every axis plot/.append style={fill}
    ]
    
    \addplot table [x index=0, y index=2] {rw_impr.csv};
    
    \addplot table [x index=0, y index=7] {rw_impr.csv};
    
    \addplot [black!15!green] table [x index=0, y index=8] {rw_impr.csv};
    
    \addplot [black] table [x index=0, y index=1] {rw_impr.csv};

    \end{axis}
\end{tikzpicture}
\caption{\label{fig:lstm_impr}Total reward percentage improvement \wrt{} no predictions (best viewed in colour).}
\end{figure*}

\subsection{LSTM and \emph{Yesterday} Predictors}

In our second set of experiments we aim at evaluating two baseline predictors, a vanilla LSTM predictor and a \emph{yesterday} predictor, which simply replicates the request of the previous day.
We measure the performance both in terms of accuracy (SMAPE) and total reward improvement. 
Given the above-discussed results, we consider a forecast horizon of 1 time step for both these predictors.

Figure~\ref{fig:smape} shows that the vanilla LSTM significantly outperforms the \emph{yesterday} predictor in terms of accuracy, achieving an average SMAPE of 6.27 \emph{vs} 11.18.
Nonetheless, these results are not reflected when measuring the improvement in terms of total reward (Figure~\ref{fig:lstm_impr}), as the vanilla LSTM is actually performing \emph{worse} than the counterpart ($+0.71\%$ \emph{vs} $+1.89\%$).
Note that our previous \emph{look-ahead} approach ($+2\%$) slightly outperforms both baseline predictors.

\section{Conclusions}

In this preliminary study concerning the integration of predictions in our online approximate algorithm for large-scale P2P-RS optimisation, we showed that, in our experiments, predicting future requests is beneficial, \emph{in the best case}, only for up to 2 time steps of forecast horizon, and that a vanilla LSTM predictor \emph{cannot} improve upon a \emph{yesterday} predictor in terms of total reward improvement, whilst achieving an almost-double accuracy.
Thus, we decisively conclude that accuracy alone cannot be taken as a sole performance measure for predictions, especially in the context of solving an online optimisation process.
Nonetheless, given our sometimes counter-intuitive findings, we will further investigate the impact of predictions on the optimisation process.
Furthermore, we aim at evaluating different and more advanced predictors, e.g., CNN-LSTM~\cite{donahue2015long}, which are specifically designed for prediction problems on inputs with spatial dependencies, like the P2P-RS problem.

\section*{Acknowledgement}

This project has received funding from the EU H2020 programme under grant agreement \#769142.


\bibliographystyle{ACM-Reference-Format}  
\bibliography{bibliography}  

\end{document}